\documentclass[preprint,12pt]{elsarticle}

\usepackage{amssymb}
\usepackage{amsmath}
\usepackage{graphicx}
\usepackage{amsmath,amssymb,amsfonts}
\usepackage{algorithm}
\usepackage{algorithmicx}
\usepackage[noend]{algpseudocode}
\usepackage{multirow}
\usepackage{breqn}
\usepackage{url}
\usepackage{ulem}
\usepackage{cleveref}
\usepackage{appendix}
\usepackage[table,xcdraw]{xcolor}
\usepackage{booktabs}
\usepackage{siunitx}
\usepackage{diagbox}
\usepackage{caption}

\newcommand{\bfphi}{\boldsymbol{\phi}}

\setcitestyle{square}

\journal{Computer Methods in Applied Mechanics and Engineering }

\begin{document}

\begin{frontmatter}



\title{Adaptive Collocation Point Strategies For Physics Informed Neural Networks via the QR Discrete Empirical Interpolation Method}

\affiliation[rice]{organization={Rice University},
            city={Houston},
            postcode={77005},
            state={TX},
            country={USA}}

\affiliation[mda]{organization={The University of Texas MD Anderson Cancer Center},
            city={Houston},
            postcode={77030},
            state={TX},
            country={USA}}

\author[rice,mda]{Adrian Celaya} 
\author[mda]{David Fuentes}
\author[rice]{Beatrice Riviere}

\begin{abstract}
Physics-informed neural networks (PINNs) have gained significant attention for solving forward and inverse problems related to partial differential equations (PDEs). While advancements in loss functions and network architectures have improved PINN accuracy, the impact of collocation point sampling on their performance remains underexplored. Fixed sampling methods, such as uniform random sampling and equispaced grids, can fail to capture critical regions with high solution gradients, limiting their effectiveness for complex PDEs. Adaptive methods, inspired by adaptive mesh refinement from traditional numerical methods, address this by dynamically updating collocation points during training but may overlook residual dynamics between updates, potentially losing valuable information. To overcome this limitation, we propose two adaptive collocation point selection strategies utilizing the QR Discrete Empirical Interpolation Method (QR-DEIM), a reduced-order modeling technique for efficiently approximating nonlinear functions. Our results on benchmark PDEs demonstrate that our QR-DEIM-based approaches improve PINN accuracy compared to existing methods, offering a promising direction for adaptive collocation point strategies.
\end{abstract}

\end{frontmatter}

\section{Introduction}
In recent years, physics-informed neural networks (PINNs) have emerged as a powerful tool for solving forward and inverse problems involving partial differential equations (PDEs) \cite{karniadakis2021physics, lu2021deepxde, raissi2019physics}. By embedding the governing PDEs directly into the neural network's loss function via automatic differentiation, PINNs provide a flexible, mesh-free alternative to traditional numerical PDE solvers like the finite difference and finite element methods. Moreover, PINNs can easily incorporate physics-based constraints and observational data into loss functions. Applications of PINNs span a diverse range of areas in computational science and engineering, including computational fluid dynamics \cite{cai2021physics}, coupled or multi-physics problems \cite{wang2023multiphysics, degen20223d}, and parameter estimation in biological systems \cite{yazdani2020systems}.

Despite their successes, tackling increasingly complex problems with PINNs poses both theoretical and practical challenges \cite{wu2023comprehensive}. Since their introduction in \cite{raissi2019physics}, numerous extensions of the original PINN framework have enhanced their computational performance and accuracy. For instance, meta-learning has led to the development of improved loss functions for PINNs \cite{psaros2022meta}, while gradient-enhanced PINNs effectively integrate information about the PDE residual's gradient into their loss function \cite{yu2022gradient}. In addition to improved loss functions, recent developments have improved network architectures for PINNs. For instance, \cite{wang2024pinn} uses neural architecture search to find optimal neural network architectures for solving certain PDEs, and \cite{lu2021physics} proposes networks that directly encode constraints like Dirichlet and periodic boundary conditions into the architecture.

PINNs minimize their loss function on a set of collocation points sampled from the problem's computational domain. The effect of these collocation points is similar to that of mesh or grid points in traditional numerical methods for PDEs. Therefore, the distribution of these points is crucial to the accuracy of PINNs. However, despite all of the recent advancements in loss functions and network architectures for PINNs, most of these methods tend to use simple collocation point sampling methods like uniformly sampling random points or an equispaced grid and do not consider the effect of collocation point sampling on their accuracy.

\subsection{Previous Work} \label{sec:previous-work}
Collocation point sampling methods can be broadly classified into two categories: fixed and adaptive. As the name suggests, fixed methods generate a set of collocation points that remains constant throughout the training process. For instance, we could use points arranged in an equispaced grid or random samples drawn from a uniform distribution over the computational domain as collocation points. While these two approaches are widespread, other fixed methods include sampling from Latin hypercubes \cite{raissi2019physics}, Sobol sequences \cite{pang2019fpinns}, and Hammersley sequences \cite{wu2023comprehensive}. 

\begin{figure}[ht!]
    \centering
    \includegraphics[width=0.7\linewidth]{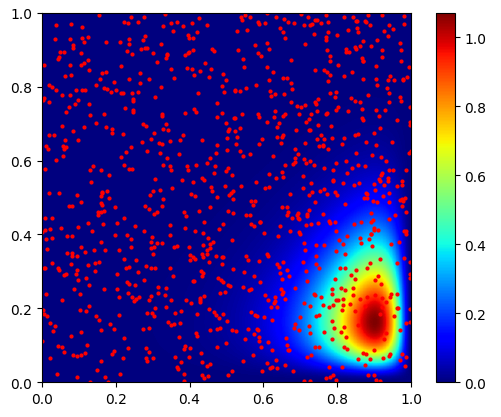}
    \caption{Visualization of $u(x, y) = 0.0005 x^2 (x - 1)^2 y^2 (y - 1)^2 \exp(10x^2 + 10y)$ with randomly sampled collocation points shown as red dots.\label{fig:intro-motivation}}
\end{figure}

An advantage of fixed sampling methods is their simplicity - they are trivial to implement and easy to use with training PINNs. However, they do not incorporate information about the residual during training and, as a result, can miss or undersample regions that contribute most to the overall error. For example, suppose the function $u(x,y) = 0.0005 x^2 (x - 1)^2 y^2 (y - 1)^2 \exp(10x^2 + 10y)$ is the solution to some PDE that we want to solve using a PINN where we sample collocation points with uniform random sampling on the domain $[0, 1] \times [0, 1]$. Figure \ref{fig:intro-motivation}  displays the solution  $u$ with the randomly sampled collocation points shown as red dots. The distribution of the collocation points highlights a common issue with fixed sampling methods: most points are in flat regions, while areas with high gradients in $u$ are undersampled. This imbalance can negatively impact the accuracy of PINNs.

While fixed sampling methods work well for simple PDEs, they may not be sufficient for more complex problems. Inspired by adaptive mesh refinement in finite element methods, \cite{lu2021deepxde} introduced the residual-based adaptive refinement method, which adds new residual points in locations with large PDE residuals - improving the accuracy of their PINN versus training with fixed sampling methods. Since its introduction, other versions of this greedy algorithm have been proposed in \cite{zeng2022adaptive, hanna2022residual, wu2023comprehensive}. Another approach for adaptive refinement is to construct a probability density function (PDF) based on the PDE residual and sample points according to this PDF. This method of sampling from a PDF instead of the points with the highest residual was first proposed in \cite{nabian2021efficient}, with similar approaches being investigated in \cite{wu2023comprehensive, zapf2022investigating, gao2023active}.

Other recent examples of adaptive collocation methods include the Retain-Resample-Release (R3) and PINN Adaptive ColLocation and Experimental (PINNACLE) schemes \cite{pmlr-v202-daw23a, lau2024pinnacle}. The R3 sampling strategy introduces a lightweight, iterative approach based on a “Retain-Resample-Release” mechanism that adaptively accumulates collocation points in high-residual regions. This method shares similarities with the greedy sampling strategy since it only keeps points that fall above a residual-based threshold. However, unlike the greedy algorithm, points whose residual falls below the threshold are replaced with new, randomly sampled points. As a result of the resample and replacement properties, the R3 method mitigates propagation failures by incrementally reinforcing under-resolved regions without the need for more collocation points. PINNACLE proposes a distribution-based sampling framework grounded in Neural Tangent Kernel (NTK) theory. It adaptively selects collocation and experimental points by maximizing a convergence criterion derived from the NTK eigenspectrum over an augmented input space. While R3 emphasizes dynamic local refinement, PINNACLE offers a global optimization strategy that balances collocation types and dynamically reallocates training point budgets during training.

A common aspect of the adaptive methods mentioned above is that they sample collocation points at fixed iteration intervals from a large set of points that is separate from the training set. As a result, these methods do not directly consider the dynamics of the residuals between these intervals, which may lead to a loss of valuable information that could enhance the convergence of the training process and produce more accurate results. To address this issue, we propose an adaptive collocation point selection strategy that employs tools from reduced-order modeling (ROM) to directly incorporate these residual dynamics into the training of PINNs for solving forward problems related to PDEs. Specifically, we turn to the QR Discrete Empirical Interpolation Method (QR-DEIM), a common technique in ROM for efficiently approximating nonlinear functions \cite{chaturantabut2010nonlinear,drmac2016new}.

To the best of our knowledge, the first application of the QR-DEIM algorithm in relation to PINNs was proposed by Forootani et al. \cite{forootani2024gs}. However, their approach focuses on inverse problems related to PDEs. Specifically, their QR-DEIM-based method is for parameter estimation, which begins with sampling points from data from the solution of a PDE with unknown parameters. In contrast, our approach is for solving the forward problem, which approximates the solution of a PDE - this divergence in focus results in fundamentally different algorithms.

\subsection{Contributions}
In this paper, we introduce two adaptive collocation point selection strategies for PINNs, both based on the QR-DEIM algorithm. The first method uses the standard QR-DEIM algorithm, while the second employs a randomized variant to improve scalability. We evaluate both approaches on four benchmark PDEs - the wave, convection, Allen–Cahn, and Burgers’ equations - and compare their performance to a wide range of fixed and adaptive sampling strategies. Our results show that both methods consistently achieve lower errors than existing methods, demonstrating their effectiveness as robust and efficient tools for adaptive sampling in PINNs.

\section{Methods} \label{sec:methods}
\subsection{Background} \label{sec:methods-background}
\subsubsection{Physics Informed Neural Networks}
Consider the following generic initial-boundary value problem defined on a space-time domain $\Omega \times [0, T]$ with solution $u$ such that
\begin{align} \label{eqn:generic-pde}
\begin{split}
    \frac{\partial u}{\partial t} &= \mathrm{L}(u) + \mathrm{F}(u), \quad \mbox{in}\quad \Omega \times (0, T], \\
    u(x, t) &= g(x, t), \quad \forall (x, t) \in \partial \Omega \times [0, T], \\
    u(x, 0) &= u_0(x), \quad \forall x \in \Omega.
\end{split}
\end{align}
Here, $\mathrm{L}$ is a linear spatial differential operator and $\mathrm{F}$ is a nonlinear function. A PINN seeks to approximate the solution $u$ by a neural network $u_{\theta}$, parameterized by weights $\theta$, such that $u_{\theta} \approx u$. The key idea behind PINNs is to train the network by minimizing a total loss function that encodes physical constraints and available data. This total loss function typically takes the form
\begin{align}
    \mathcal{L}_{\text{total}}(u_{\theta}) = \mathcal{L}_{\text{residual}}(u_{\theta}) + 
    \mathcal{L}_{\text{IC}}(u_{\theta}) +
    \mathcal{L}_{\text{boundary}}(u_{\theta}).
\end{align}

The first constituent loss function $\mathcal{L}_{\text{residual}}$ is computed over a set of collocation points $\mathcal{T}_{\text{residual}} \subset \Omega \times (0, T]$ as
\begin{align}\label{eqn:pde-loss}
    \mathcal{L}_{\text{residual}}(u_{\theta}) = \frac{1}{\vert \mathcal{T}_{\text{residual}}\vert} \sum_{(x_i, t_i) \in \mathcal{T}_{\text{residual}}} r(x_i, t_i)^2,
\end{align}
where the pointwise residual $r(x, t)$ is defined as
\begin{align}\label{eqn:pde-residual}
    r(x, t) = \frac{\partial u_{\theta}(x, t)}{\partial t} - \mathrm{L}\left(u_{\theta}\left(x, t\right)\right) - \mathrm{F}\left(u_{\theta}\left(x, t\right)\right).
\end{align}
Automatic differentiation is used to compute the partial derivatives for the terms $\frac{\partial u_{\theta}}{\partial t}$ and $\mathrm{L}(u_{\theta})$. The other terms in $\mathcal{L}_{\text{total}}$ are defined as
\begin{align}
    \mathcal{L_{\text{IC}}}(u_{\theta}) = \frac{1}{\vert \mathcal{T}_{\text{IC}} \vert} \sum_{x_i \in \mathcal{T}_{\text{IC}}} \left( u_{\theta}(x_i, 0) - u_0(x_i) \right)^2
\end{align}
and
\begin{align}
    \mathcal{L_{\text{boundary}}}(u_{\theta}) = \frac{1}{\vert \mathcal{T}_{\text{boundary}} \vert} \sum_{(x_i, t_i) \in \mathcal{T}_{\text{boundary}}} \left( u_{\theta}(x_i, t_i) - g(x_i, t_i) \right)^2,
\end{align}
where $\mathcal{T}_{\text{IC}}\subset \Omega$, $\mathcal{T}_{\text{boundary}} \subset \partial\Omega \times [0, T]$ are sets of collocation points for the initial and boundary conditions, respectively.

Depending on the PDE, it is possible to strongly enforce the initial and boundary conditions by incorporating them directly into the architecture of the neural network through output transformations that guarantee satisfaction. For instance, consider a one-dimensional PDE defined on the spatial domain $\Omega = (-1, 1)$ with homogeneous Dirichlet boundary conditions and an initial condition $u_0$. One can define a transformed output $\hat{u}_{\theta}$ such that
\begin{align}\label{eqn:strongly-enforce}
    \hat{u}_{\theta}(x, t) = t (x + 1) (x - 1)u_{\theta}(x, t) + u_0(x), 
\end{align}
which guarantees that $\hat{u}_{\theta}(x, 0) = u_0(x)$ and $\hat{u}_{\theta}(-1, t) =\hat{u}_{\theta}(1, t) = 0$ in the particular case where the initial condition happens to satisfy the boundary condition. While this is not true in general, this simplifying assumption is done to focus the proposed methodology on the selection of interior points.  We also note that the transform $\hat{u}_\theta$ is applied to all the methods we test in this paper. This technique is particularly useful when the exact form of the initial and boundary conditions is known and easy to incorporate, reducing the need to approximate them via auxiliary loss terms. However, in many cases - especially for complex or data-driven conditions - they are instead enforced weakly via the loss terms $\mathcal{L}_{\text{IC}}$ and $\mathcal{L}_{\text{boundary}}$.

\begin{figure}[ht!]
    \centering
    \includegraphics[width=\linewidth]{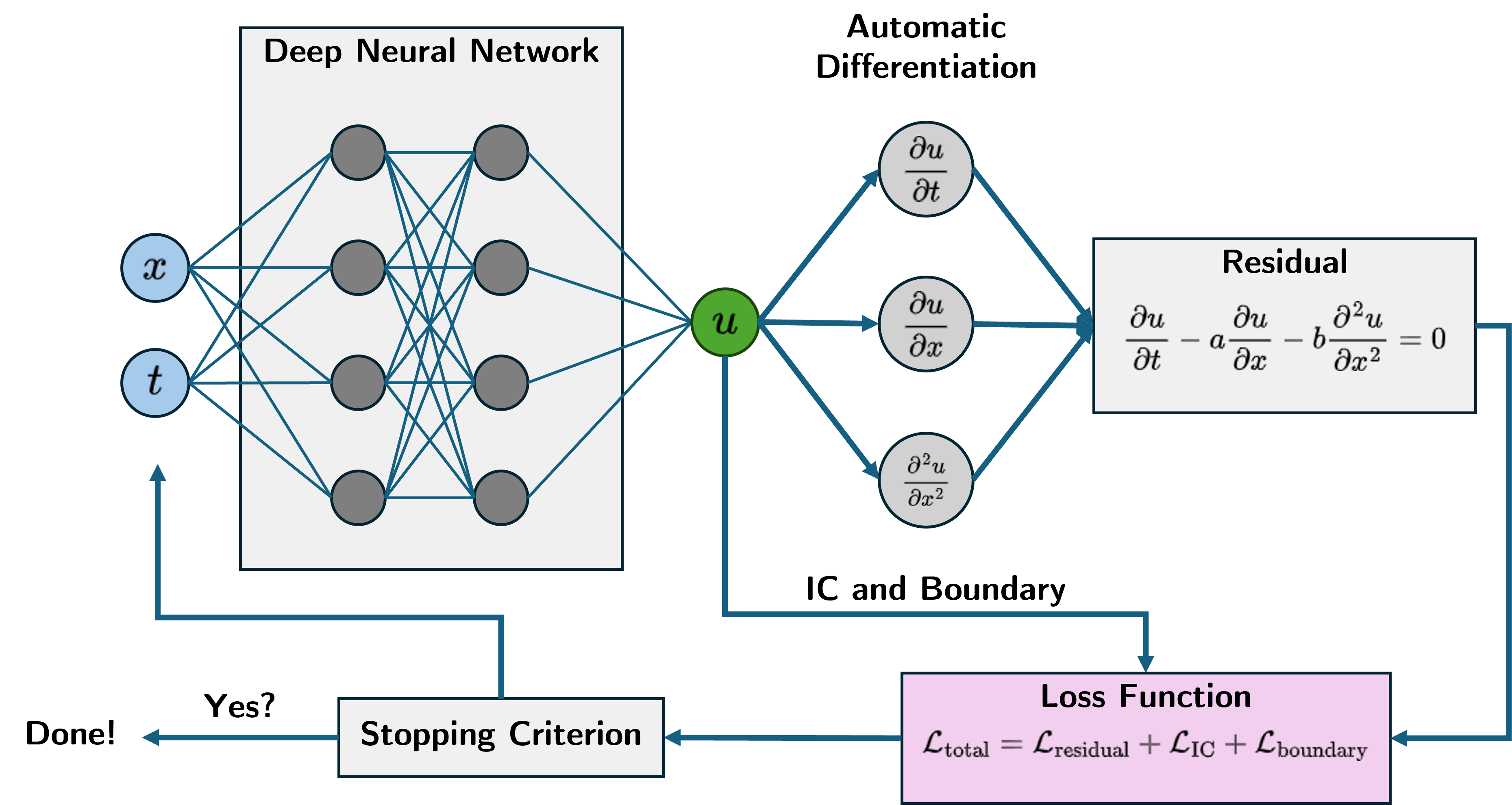}
    \caption{Visualization of a typical PINN training loop.\label{fig:pinn-framework}}
\end{figure}

The training process involves minimizing the total loss $\mathcal{L}_{\text{total}}$ with respect to the network parameters $\theta$ using stochastic optimization (i.e., Adam or L-BFGS). Training is usually terminated either upon reaching a maximum number of iterations or when the total loss falls below a predefined threshold. Figure \ref{fig:pinn-framework} illustrates a typical PINN training loop.

\subsubsection{The QR Discrete Empirical Interpolation Method}
ROM seeks to reduce the computational complexity and time of large-scale dynamical systems by finding lower-dimensional approximations that can replicate similar input-output responses. An area where ROM is particularly advantageous is high-dimensional systems of ordinary differential equations resulting from discretizing PDEs. A numerical discretization (i.e., finite difference) of (\ref{eqn:generic-pde}) results in
\begin{align} \label{eqn:ode-system}
    \frac{d}{dt}\mathbf{u}(t) = \mathbf{A}\mathbf{u}(t) + \mathbf{F}(\mathbf{u}(t)),
\end{align}
where $\mathbf{u}(t) = [u_1(t), \dots, u_n(t)]^T \in \mathbb{R}^n$, the constant matrix $\mathbf{A} \in \mathbb{R}^{n \times n}$ is a discrete approximation of $\mathrm{L}$, and $\mathbf{F}$ is the nonlinear function $\mathrm{F}$ evaluated at $\mathbf{u}(t)$ component-wise, i.e., $\mathbf{F}(\mathbf{u}(t))  = [\mathrm{F}(u_1(t)), \dots, \mathrm{F}(u_n(t))]^T \in \mathbb{R}^n$.

Projection-based techniques are common in ROM \cite{lucia2003projection, amsallem2012stabilization, wang2002projection}. These methods construct a much smaller system of order $k \ll n$ that approximates the original system (i.e., (\ref{eqn:ode-system})) using a subspace spanned by a reduced basis of dimension $k$ in $\mathbb{R}^n$. In particular, let $\mathbf{V}_k \in \mathbb{R}^{n \times k}$ be a matrix whose columns form an orthonormal basis of this reduced space. By replacing $\mathbf{u}(t)$ with $\mathbf{V}_k\hat{\mathbf{u}}(t)$, $\hat{\mathbf{u}}(t) \in \mathbb{R}^k$, and projecting (\ref{eqn:ode-system}) onto the subspace spanned by the columns of $\mathbf{V}_k$, the reduced order system is of the form:
\begin{align}
    \frac{d}{dt}\hat{\mathbf{u}}(t) = \mathbf{V}_k^T\mathbf{A}\mathbf{V}_k\hat{\mathbf{u}}(t) + \mathbf{V}_k^T\mathbf{F}(\mathbf{V}_k\hat{\mathbf{u}}(t)).
\end{align}

The selection of a reduced order basis significantly influences the approximation's quality. Techniques for constructing reduced bases rely on the observation that the solution space for a given system is often embedded in a low-dimensional manifold. Consequently, the reduced basis is typically tailored to the specific problem. These methods create global basis functions derived from \textit{snapshots} - discrete samples of trajectories from a set of different inputs and boundary conditions from the full-order system.

Proper orthogonal decomposition (POD) is a common method for constructing a low-dimensional approximation representation of a subspace in Hilbert space. Given a set of snapshots $\mathcal{U} = \{\mathbf{u}_1, \dots, \mathbf{u}_c\} \subset \mathbb{R}^n$ such that $r = \text{rank}(\text{span}(\mathcal{U}))$, a POD basis of rank $k < r$ is the set of orthonormal vectors $\{\bfphi_1, \dots, \bfphi_k\}$ whose span best approximates the space $\text{span}(\mathcal{U})$. In other words, the vectors $\{\bfphi_1, \dots, \bfphi_k\}$ are the solution to
\begin{align} \label{eqn:basis-opt}
    \min_{\bfphi_1, \dots \bfphi_k} \sum_{i=1}^c \Vert \mathbf{u}_i - \sum_{j=1}^k(\mathbf{u}_j^T\bfphi_i)\bfphi_j \Vert _2^2.
\end{align}

It is well known that the solution (\ref{eqn:basis-opt}) is given by the singular value decomposition (SVD) of the matrix $\mathbf{U} = [\mathbf{u}_1| \dots| \mathbf{u}_r] \in \mathbb{R}^{n \times r}$, which we call a \textit{snapshot matrix}. Let $\mathbf{U} = \mathbf{V} \boldsymbol{\Sigma} \mathbf{W}^T$ be the SVD of $\mathbf{U}$, where $\mathbf{V} = [\mathbf{v}_1| \dots| \mathbf{v}_r] \in \mathbb{R}^{n \times r}$ and $\mathbf{W} = [\mathbf{w}_1| \dots| \mathbf{w}_r] \in \mathbb{R}^{c \times r}$ are orthogonal and $\boldsymbol{\Sigma} = \text{diag}(\sigma_1, \dots, \sigma_r) \in \mathbb{R}^{r \times r}$ with $\sigma_1 \geq \sigma_2 \geq \dots \geq \sigma_r > 0$. The POD basis is then chosen as $\{\mathbf{v}_1, \dots, \mathbf{v}_k\}$.

Although POD has been shown to be successful in providing reduced-order models for numerous applications such as compressible flow \cite{bui2003proper}, fluid dynamics \cite{berkooz1993proper}, and optimal control \cite{ravindran2000reduced, gubisch2017proper}, its computational cost is high  \cite{kramer2019nonlinear}. Indeed, while POD significantly reduces the dimensionality of the system, evaluating the nonlinear term $\mathbf{F}(\mathbf{V}_k \hat{\mathbf{u}}(t))$ still requires computations in the high-dimensional ambient space $\mathbb{R}^n$, which can become a bottleneck for large-scale systems.

The Discrete Empirical Interpolation Method (DEIM) is an efficient way to overcome the difficulty described above \cite{chaturantabut2009discrete, chaturantabut2010nonlinear}. This method approximates a high-dimensional nonlinear function $\mathbf{f}: \mathbb{R}^n \to \mathbb{R}^n$ by projecting it onto a subspace that approximates the space of the nonlinear function and is spanned by a basis of dimension $m \ll n$. In this case, let $\mathbf{f}(t) = \mathbf{F}(\mathbf{V}_k\hat{\mathbf{u}}(t))$. The DEIM approximation from projecting $\mathbf{f}(t)$ onto the subspace spanned by the basis $\{\mathbf{y}_1, \dots \mathbf{y}_m\} \subset \mathbb{R}^n$ is of the form
\begin{align}
    \mathbf{f}(t) \approx \mathbf{Y}\mathbf{c}(t),
\end{align}
where $\mathbf{Y} = [ \mathbf{y}_1| \dots| \mathbf{y}_m ] \in \mathbb{R}^{n \times m}$ and $\mathbf{c}(t) \in \mathbb{R}^m$ is a coefficient vector. To determine $\mathbf{c}(t)$, DEIM selects $m$ rows from the linear system $\mathbf{f}(t) = \mathbf{Y}\mathbf{c}(t)$. In particular, let $\mathbf{\Pi} = [ \mathbf{e}_{\pi_1}| \dots| \mathbf{e}_{\pi_m} ] \in \mathbb{R}^{n \times m}$ be a permutation matrix, where $\mathbf{e}_{j}$ is the $j$-th column of the identity matrix. If the matrix $\mathbf{\Pi}^T\mathbf{Y}$ is non-singular, then $\mathbf{c(t)} = \left( \mathbf{\Pi}^T\mathbf{Y} \right)^{-1}\mathbf{\Pi}^T\mathbf{f}(t)$ and our final DEIM approximation is given by
\begin{align}
    \mathbf{f}(t) \approx \mathbf{Y}\left( \mathbf{\Pi}^T \mathbf{Y} \right)^{-1}\mathbf{\Pi}^T\mathbf{f}(t).
\end{align}

To obtain this approximation, DEIM requires the projection basis (i.e, $\{ \mathbf{y}_1, \dots, \mathbf{y}_m \}$) and the interpolation indices (i.e., $\{ \pi_1, \dots, \pi_m \}$). To determine the projection basis, DEIM applies POD to the nonlinear snapshots of the full-order system (i.e., $\{ \mathbf{F}(\mathbf{u}(t_1)), \dots, \mathbf{F}(\mathbf{u}(t_{n_s})) \}$). DEIM selects the $m$ interpolation indices in a greedy way such that the matrix $\mathbf{\Pi}^T\mathbf{Y}$ is non-singular and the indices $\{\pi_1, \dots, \pi_m \}$ are hierarchical and non-repeating.

To improve the numerical stability and robustness of the index selection, the QR-DEIM method was proposed as a DEIM variant that replaces greedy index selection with a pivoted QR factorization \cite{drmac2016new}. More specifically, QR-DEIM applies a QR decomposition with column pivoting to $\mathbf{Y}^T$ so that
\begin{align}
    \mathbf{Y}^T\mathbf{\Pi} = \mathbf{Q}\mathbf{R},
\end{align}
where $\mathbf{Q} \in \mathbb{R}^{m \times m}$ is orthogonal, $\mathbf{R} \in \mathbb{R}^{m \times n}$ is upper triangular. The selected interpolation indices correspond to the rows that contain non-zero values in the first $m$ columns of $\mathbf{\Pi}$.

\subsection{Adaptive Collocation Points via the QR-DEIM} \label{sec:methods-qr-deim}
We propose a strategy for adaptive collocation point selection in PINNs inspired by the QR-DEIM method. This approach involves constructing a snapshot matrix based on the residuals obtained from a set of snapshot points throughout the training process. Subsequently, we perform a singular value decomposition of this snapshot matrix and apply QR decomposition with column pivoting to select a subset of the snapshot points that we add to the training data. With this process, we can encode the residual dynamics between updates to the training set and use this information to select new collocation points. In addition to adding new points, we introduce a convergence degree criterion to remove training points that have already been resolved by the network.

Again, let $u_{\theta}$ be a neural network with trainable parameters $\theta$. For the remainder of this work, we will assume that we strongly enforce initial and boundary conditions as described in (\ref{eqn:strongly-enforce}). As a result, our loss function reduces to $\mathcal{L}_{\text{residual}}$. Let $\mathcal{T} = \mathcal{T}_{\text{residual}} = \{ (x_i, t_i) \}_{i=1}^{N_{\text{train}}}$ be the set of collocation points with which we train $u_{\theta}$. Additionally, we define the set $\mathcal{S} = \{ (x_i, t_i) \}_{i=1}^{N_{\text{snapshot}}}$, with $\mathcal{S} \cap \mathcal{T} = \emptyset$, as a set of snapshot points that are sampled separately from the training points.

Our collocation point selection strategy begins with $N_{\text{train}}$ randomly sampled points. For a predefined period of $P$ iterations, we compute the residuals over the snapshot set $\mathcal{S}$ as 
\begin{align}
\mathbf{r}_j(\mathcal{S}) = [ r(x_1, t_1), \dots, r(x_{N_{\text{snapshots}}}, t_{N_{\text{snapshots}}}) ]^T \in \mathbb{R}^{N_{\text{snapshots}}},
\end{align}
for $j = 1, \dots, P$. At the end of every $P$-th iteration, we construct a residual snapshot matrix $\mathbf{R} = [\mathbf{r}_1(\mathcal{S}), \dots, \mathbf{r}_{P}(\mathcal{S})] \in \mathbb{R}^{N_{\text{snapshots}} \times P}$ and compute the SVD of $\mathbf{R}$ as $\mathbf{R} = \mathbf{V}\boldsymbol{\Sigma}\mathbf{W}^T$. The singular values in $\boldsymbol{\Sigma}$ represent the significance of their corresponding modes in the residual snapshot matrix. To best capture these dominant features in the residual dynamics, we select the first $k$ columns of $\mathbf{V}$ such that
\begin{align}\label{eqn:find-k}
    k = \min 
    \left\{ i = 1, \dots, P \mid 1 - \frac{\sum_{j=1}^i \sigma_j}{\text{Tr}(\boldsymbol{\Sigma})} \leq \varepsilon \right\},
\end{align}
where $\varepsilon$ is a user-defined energy threshold. This formulation ensures that the selected subspace captures at least $(1 - \varepsilon) \times 100$\% of the total snapshot residual energy. This selection yields the reduced matrix $\mathbf{V}_k \in \mathbb{R}^{N_{\text{snapshots}} \times k}$, which forms the basis for selecting new collocation points in the next stage of the update procedure.

To select new training points from the snapshot set $\mathcal{S}$, we apply QR decomposition with column pivoting to $\mathbf{V}_k$ so that $\mathbf{V}_k^T\boldsymbol{\Pi} = \mathbf{Q}\mathbf{R}$. Like with QR-DEIM, we select the indices that correspond to the rows containing non-zero values from the first $k$ columns of $\boldsymbol{\Pi}$. In other words, we define the index selection set as
\begin{align}
    \mathcal{I} = \{ i \mid \exists \text{ } j_0 \in \{1, \dots, k \} \text{ such that } \mathbf{\Pi}_{ij_0} = 1 \}
\end{align}
and add $\{ (x_i, t_i) \in \mathcal{S} : i \in \mathcal{I} \}$ to the training set $\mathcal{T}$. This selection yields $k$ new points to add to our training set $\mathcal{T}$ from our snapshot set $\mathcal{S}$.

To ensure that the training set remains fixed in size and concentrated on under-resolved regions of the computational domain, we incorporate a pruning mechanism that removes collocation points that have already been well-resolved by the network. Specifically, we track the magnitude of the residuals at each training point $(x_i, t_i) \in \mathcal{T}$ both at the end of the previous update and just before the current update. We denote these residual magnitude vectors as $\hat{\mathbf{r}}_{\text{old}}(\mathcal{T})$ and $\hat{\mathbf{r}}_{\text{new}}(\mathcal{T})$, respectively. To quantify pointwise convergence, we define a convergence degree vector as
 \begin{align}
 \mathbf{d} = \log_2 \left( \frac{\hat{\mathbf{r}}_{\text{old}}(\mathcal{T})}{\hat{\mathbf{r}}_{\text{new}}(\mathcal{T})} \right) \in \mathbb{R}^{N_{\text{train}}},
 \end{align}
where the logarithm and division operators are applied element-wise. Each entry of $\mathbf{d}$ reflects the rate at which the residual at a point has decreased - larger values indicate more significant local convergence. After determining the number of new points to add, $k$, as described in (\ref{eqn:find-k}), we prune the $k$ training points corresponding to the largest entries in $\mathbf{d}$. However, to prevent the removal of points that are either stagnating or diverging, we introduce a safeguard. If any of the top $k$ values in $\mathbf{d}$ are negative, indicating an increase in the residual, we skip the update and retain the current training set. This check ensures that the pruning procedure only proceeds when there is clear evidence of local convergence at the selected points.

The final step in this update is to resample our snapshot points $\mathcal{S}$. We resample these points to ensure continued exploration of the domain, which helps to prevent bias toward specific regions and allows us to capture a broader range of error dynamics. Furthermore, this resampling helps maintain the size of $N_{\text{snapshots}}$ at a manageable level, which keeps the computational cost of computing the SVD and QR decomposition within reasonable limits. 

The update process described above repeats until the PINN satisfies a predefined stopping criterion (i.e., maximum number of iterations). Algorithm~\ref{alg:qr-deim-collocation} provides a complete description of the update procedure.

\begin{algorithm}[H]
\caption{
QR-DEIM Collocation Point Update \\
\textit{Input}: Residual snapshot matrix $\mathbf{R}$, training set $\mathcal{T}$, snapshot set $\mathcal{S}$, previous residual magnitudes $\hat{\mathbf{r}}_{\text{old}}(\mathcal{T})$, residual function $r(x, t)$, energy threshold $\varepsilon$
}
\label{alg:qr-deim-collocation}
\begin{algorithmic}[1]
\vspace{0.5em}
\State Compute SVD: $\mathbf{R} = \mathbf{V} \boldsymbol{\Sigma} \mathbf{W}^T$ \vspace{0.5em}
\State Compute $k$: $k \gets \min \left\{ i = 1, \dots, P \mid 1 - \frac{\sum_{j=1}^{i} \sigma_j}{\text{Tr}(\boldsymbol{\Sigma})} \leq \varepsilon \right\}$ \vspace{0.5em}

\State Get first $k$ columns of $\mathbf{V}$: $\mathbf{V}_k \gets \mathbf{V}[:, 1, \dots, k]$ \vspace{0.5em}
\State Compute QR with column pivoting: $\mathbf{V}_k^T \boldsymbol{\Pi} = \mathbf{Q} \mathbf{R}$ \vspace{0.5em}
\State Get selection indices: $\mathcal{I} = \{ i \mid \exists \text{ } j_0 \in \{1, \dots, k \} \text{ such that } \mathbf{\Pi}_{ij_0} = 1 \}$ \vspace{0.5em}

\State Get residual magnitudes: $\hat{\mathbf{r}}_{\text{new}}(\mathcal{T}) \gets [|r(x_1, t_1)|, \dots, |r(x_{N_{\text{train}}}, t_{N_{\text{train}}})|]^T$ \vspace{0.5em}
\State Compute convergence degrees: $\mathbf{d} \gets \log_2\left( \frac{\hat{\mathbf{r}}_{\text{old}}(\mathcal{T})}{\hat{\mathbf{r}}_{\text{new}}(\mathcal{T})} \right)$ \vspace{0.5em}
\If{Any of the top $k$ values of $\mathbf{d}$ are negative,} \vspace{0.5em}
    \State Leave $\mathcal{T}$ unchanged. \vspace{0.5em}
\Else \vspace{0.5em}
    \State $\mathcal{J} \gets$ Indices of top $k$ values in $\mathbf{d}$ \vspace{0.5em}
    \State Remove converged points: $\mathcal{T} \gets \mathcal{T} \setminus \{ (x_j, t_j) \in \mathcal{T} \mid j \in \mathcal{J} \}$ \vspace{0.5em}
    \State Add new points: $\mathcal{T} \gets \mathcal{T} \cup \{ (x_i, t_i) \in \mathcal{S} \mid i \in \mathcal{I} \}$ \vspace{0.5em}
\EndIf

\State Resample snapshot set $\mathcal{S}$ \vspace{0.5em}
\State Update training residuals: $\hat{\mathbf{r}}_{\text{old}}(\mathcal{T}) \gets \hat{\mathbf{r}}_{\text{new}}(\mathcal{T})$
\end{algorithmic}
\end{algorithm}



Because the computational cost of the SVD and QR with column pivoting can be high - particularly as the size of the snapshot set and number of residual evaluations increase - we propose a randomized variant of our QR-DEIM-based sampling method, referred to as QR-DEIM-R. This variant starts by replacing the full SVD in Step~1 of Algorithm~\ref{alg:qr-deim-collocation} with the randomized SVD introduced in~\cite{halko2011finding}. The randomized approach approximates the leading singular vectors by projecting the snapshot matrix onto a lower-dimensional subspace using random projections, substantially reducing the computational cost of the decomposition. Whereas the full SVD of an $m \times n$ dense matrix requires $\mathcal{O}(mn^2)$ operations (assuming $m \geq n$), the randomized SVD achieves comparable accuracy at a reduced cost of approximately $\mathcal{O}(mn \log k)$, where $k \ll n$ is the target rank.

To mitigate the cost of the QR decomposition with column pivoting, we note that in Algorithm~\ref{alg:qr-deim-collocation}, the matrices $\mathbf{Q}$ and $\mathbf{R}$ are not used. Instead, our primary focus is on the pivot indices derived from the  matrix $\mathbf{\Pi}$. Hence, we again turn to randomized methods to approximate the behavior of the pivot selection from the full QR decomposition with column pivoting. Given the matrix $\mathbf{V}_k$ from the randomized SVD described above, we begin this approximation by computing the norms of the columns of $\mathbf{V}_k^T$ and normalize by their sum to create a discrete probability distribution $\mathbf{p} \in \mathbb{R}^{N_{\text{snapshots}}}$. Next, we randomly sample $\ell = k + z$ columns indexed by $\{ j_1, \dots, j_{\ell} \}$ according to the probabilities defined in $\mathbf{p}$. Here, $z \geq 0$ is an oversampling parameter. After obtaining the sampled columns, we apply the QR decomposition with column pivoting to the reduced matrix $\mathbf{C} \in \mathbb{R}^{k \times \ell}$, such that $\mathbf{C} \mathbf{\Pi} = \mathbf{Q}\mathbf{R}$. We then select our new index set as follows:
\begin{align}\label{eqn:randomized-index}
    \mathcal{I} = \{ j_a \mid \exists \text{ } b \in \{1, \dots k\} \text{ such that } \mathbf{\Pi}_{ab} = 1 \}.
\end{align}

In QR-DEIM-R, we specify $k$ directly as an input in place of the energy tolerance $\varepsilon$. We then replace Steps~1 through~3 of Algorithm~\ref{alg:qr-deim-collocation} with the randomized SVD, computing an approximate factorization $\mathbf{R} \approx \mathbf{V}_k \boldsymbol{\Sigma}_k \mathbf{W}_k^T$. Steps~4 and~5 are replaced by the index selection described in~(\ref{eqn:randomized-index}). The remaining steps of the algorithm proceed as in the full QR-DEIM procedure, starting with Step~6. Algorithm~\ref{alg:qr-deim-r-collocation} provides a complete description of the update procedure.

\begin{algorithm}[H]
\caption{
Randomized QR-DEIM Collocation Point Update \\
\textit{Input}: Residual snapshot matrix $\mathbf{R}$, training set $\mathcal{T}$, snapshot set $\mathcal{S}$, previous residual magnitudes $\hat{\mathbf{r}}_{\text{old}}(\mathcal{T})$, residual function $r(x, t)$, target rank $k$, oversampling parameter $z$
}
\label{alg:qr-deim-r-collocation}
\begin{algorithmic}[1]
\vspace{0.5em}
\State Compute randomized SVD: $\mathbf{R} \approx \mathbf{V}_k \boldsymbol{\Sigma}_k \mathbf{W}_k^T$ \vspace{0.5em}
\State Compute column norms: $c_j = \|\mathbf{V}_k^T[:, j]\|_2^2$ for $j = 1, \dots, N_{\text{snapshots}}$
\vspace{0.5em}
\State Normalize probabilities: $\mathbf{p} = \frac{1}{\sum_i c_i} [c_1, \dots, c_{N_{\text{snapshots}}} ]^T$
\vspace{0.5em}
\State Randomly sample $\ell = k + z$ columns $\{ j_1, \dots, j_{\ell}\}$ according to $\mathbf{p}$
\vspace{0.5em}
\State Form reduced matrix: $\mathbf{C} = \mathbf{V}_k^T[:, j_1, \dots, j_{\ell}]$
\vspace{0.5em}
\State Compute QR: $\mathbf{C} \mathbf{\Pi} = \mathbf{Q} \mathbf{R}$
\vspace{0.5em}
\State Get selection indices: $\mathcal{I} = \{ j_a \mid \exists \text{ } b \in \{1, \dots k\} \text{ such that } \mathbf{\Pi}_{ab} = 1 \}$
\vspace{0.5em}
\State Continue Algorithm~\ref{alg:qr-deim-collocation} from Step~6
\end{algorithmic}
\end{algorithm}

\subsection{Training \& Testing Protocols} \label{sec:methods-protocols}
To assess the effectiveness of our proposed QR-DEIM-based sampling strategy, we evaluate it using the wave, convection, Allen-Cahn, and Burgers' equations and compare its performance against the following methods: uniform random sampling (non-adaptive), Hammersley sampling (non-adaptive), uniform random sampling with random resampling (random-r, non-adaptive), residual-based adaptive refinement with greed (RAR-G) \cite{lu2021deepxde, wu2023comprehensive}, residual-based adaptive distribution (RAD) \cite{wu2023comprehensive}, residual-based adaptive refinement with distribution (RAR-D) \cite{wu2023comprehensive}, R3 \cite{pmlr-v202-daw23a}, and PINNACLE \cite{lau2024pinnacle}.

We use 2,000 collocation points for all of our benchmark PDEs and the recommended default hyperparameters for each of the adaptive sampling methods. Table~\ref{tab:hyperparameters} summarizes these hyperparameters. For the non-adaptive random resampling method, we resample a new training set every 1,000 iterations. For our QR-DEIM-based approach, the default parameters are a snapshot set of 1,000 points, a snapshot period of 1,000 iterations, and an energy threshold equal to 0.005. For the randomized QR-DEIM variant, we use the same number of snapshot points and snapshots, but set the default target rank $k$ equal to 100 and set the oversampling parameter $z$ equal to 50.

To isolate the effects of the sampling strategies, we intentionally adopt a consistent and straightforward training setup across all experiments. This setup includes using a relatively simple network architecture and a standard first-order optimizer with a fixed learning rate schedule. While more advanced architectures or second-order optimization schemes could improve overall accuracy or convergence, our goal in this work is not necessarily to push the state-of-the-art performance for individual benchmark problems. Instead, we seek to compare sampling methods under controlled, minimal training pipelines. This design ensures that any observed improvements are attributable to the sampling strategy rather than confounding architectural or training effects.

For all experiments, we use a feedforward, fully connected neural network with five hidden layers and 64 hidden units per hidden layer. We use the tanh activation function between each layer. The output of each network is modified to strongly enforce the initial and boundary conditions for each benchmark problem. We use the Adam optimizer with a cosine annealing learning rate schedule. The initial learning rate for our schedule is equal to 0.001. The stopping criterion for training is a maximum number of iterations, which we set equal to 100,000 iterations for all problems except for the convection equation, where we use 1,000,000 iterations. This increased number of iterations is necessary because the convection equation produces a particularly ill-conditioned loss landscape, making convergence of first-order optimizers slower \cite{rathore2024challenges}. Our loss function is the mean squared error of the PDE residual on the training set (see (\ref{eqn:pde-loss}) and (\ref{eqn:pde-residual})). All networks use a validation set consisting of the same 10,000 randomly sampled points, and we save the weights of each model that results in the lowest validation loss.

To evaluate the accuracy of our solutions, we use the relative $\ell_2$ error between the true solution $u$ and the predicted solution $u_{\theta}$ evaluated at a set of test points $\left\{ (x_1, t_1), \dots, (x_{N_{\text{test}}}, t_{N_{\text{test}}}) \right\}$. In other words, let $\mathbf{u}_{\text{true}} = [u(x_1, t_1), \dots, u(x_{N_{\text{test}}}, t_{N_{\text{test}}})]^T$ and $\mathbf{u}_{\text{pred}} = [u_{\theta}(x_1, t_1), \dots, u_{\theta}(x_{N_{\text{test}}}, t_{N_{\text{test}}})]^T$. The relative $\ell_2$ error between these two solution vectors is defined as 
\begin{align}
\frac{\Vert \mathbf{u}_{\text{true}} - \mathbf{u}_{\text{pred}}\Vert_2}{\Vert \mathbf{u}_{\text{true}} \Vert_2}.
\end{align}

Because all the collocation point methods we assess are at least partially stochastic, we test each method on every benchmark problem 10 times and report the mean and standard deviation of the relative $\ell_2$ error on the same test set which are the points in a $256\times 100$ grid in the spatial-temporal domain $[-1, 1] \times [0, 1]$.

All of our models and experiments are implemented in Python using PyTorch~2.7.1. All experiments use an Nvidia RTX A4000 GPU and Intel Xeon w7-3445 CPU.

\begin{table}[ht!]
\centering
\bgroup
\def\arraystretch{1.25}
\resizebox{\textwidth}{!}{%
\begin{tabular}{lllll}
\hline
\multicolumn{1}{l}{Method} & \multicolumn{1}{l}{Initial Points} & \multicolumn{1}{l}{Sampling Interval} & \multicolumn{1}{l}{Points per Update} & \multicolumn{1}{l}{Implementation Details} \\ \hline
\rowcolor[HTML]{EFEFEF} 
RAR-G \cite{wu2023comprehensive} & 1,000 & 1,000 & 10 & Candidate pool size of 10,000 \\
RAR-D \cite{wu2023comprehensive} & 1,000 & 1,000 & 10 & \begin{tabular}[c]{@{}l@{}}$k=2$, $c=0$, candidate pool\\ size of 10,000\end{tabular} \\
\rowcolor[HTML]{EFEFEF} 
RAD \cite{wu2023comprehensive} & 2,000 & 1,000 & \begin{tabular}[c]{@{}l@{}}Release and\\ resample all points\end{tabular} & \begin{tabular}[c]{@{}l@{}}$k=2$, $c=0$; candidate pool\\ size of 10,000\end{tabular} \\ 
R3 \cite{pmlr-v202-daw23a} & 2,000 & 1 & \begin{tabular}[c]{@{}l@{}}Release and\\ resample $k$\\ points\end{tabular} & \begin{tabular}[c]{@{}l@{}}Residual threshold is mean of\\ the residual magnitudes,\\ $k$ depends on threshold.\end{tabular} \\
\rowcolor[HTML]{EFEFEF} 
PINNACLE \cite{lau2024pinnacle} & 1,000 & 1,000 & 100 & \begin{tabular}[c]{@{}l@{}}$\mathcal{Z}_{\text{pool}} = 4000$,\\ $\mathcal{Z}_{\text{ref}} = 500$,\\ $m = 100$ NTK modes\end{tabular} \\
QR-DEIM & 2,000 & 1,000 & \begin{tabular}[c]{@{}l@{}}Release and\\ resample $k$\\ points\end{tabular} & \begin{tabular}[c]{@{}l@{}}Candidate pool size (i.e., \\ snapshot set) of 1,000,\\ $\varepsilon=0.005$,\\ $k$ depends on $\varepsilon$\end{tabular} \\
\rowcolor[HTML]{EFEFEF}
QR-DEIM-R & 2,000 & 1,000 & \begin{tabular}[c]{@{}l@{}}Release and\\ resample 100\\ points\end{tabular} & \begin{tabular}[c]{@{}l@{}}Candidate pool size (i.e., \\ snapshot set) of 1,000, \\ $z=50$\end{tabular} \\ \hline
\end{tabular}%
}
\caption{Summary of adaptive sampling default hyperparameters. The variables for the implementation details are defined in \cite{wu2023comprehensive,pmlr-v202-daw23a,lau2024pinnacle} for the published methods. \label{tab:hyperparameters}}
\egroup
\end{table}

\section{Results \& Discussion} \label{sec:results}
In this section, we evaluate the accuracy of the sampling methods introduced in Sections \ref{sec:previous-work} and \ref{sec:methods-qr-deim} using the training and testing protocols described in Section \ref{sec:methods-protocols}. Our experiments focus on the wave, convection, Allen–Cahn, and Burgers’ equations. Additionally, we present ablation studies examining key parameters of our QR-DEIM-based approaches and discuss the computational costs and considerations for our methods.

\subsection{Wave Equation}
We begin by examining the wave equation stated below:
\begin{align*}
\begin{split}
    \frac{\partial^2u}{\partial t^2} &= 4\frac{\partial^2u}{\partial x^2}, \quad \mbox{in} \quad (-1, 1) \times (0, 1], \\
    u(-1, t) &= u(1, t) = 0, \quad \forall t \in [0, 1], \\  
    u(x, 0) &= \sin(\pi x) + \frac{1}{2}\sin(4\pi x), \quad \forall x \in (-1, 1),\\
     \frac{\partial u}{\partial t}(x,0) &= 0, \quad \forall x \in (-1, 1). 
\end{split}
\end{align*}
The true solution exhibits multiscale behavior in both the spatial and temporal directions, making this an effective benchmark for assessing how well different collocation strategies resolve global structure and local oscillations. Table~\ref{tab:results-accuracy} reports the mean and standard deviation of the relative $\ell_2$ error across ten training runs for each of the sampling methods that we test. Both the QR-DEIM and QR-DEIM-R methods achieve lower average errors than the other approaches.

\begin{figure}[ht!]
    \centering
    \includegraphics[width=0.85\linewidth]{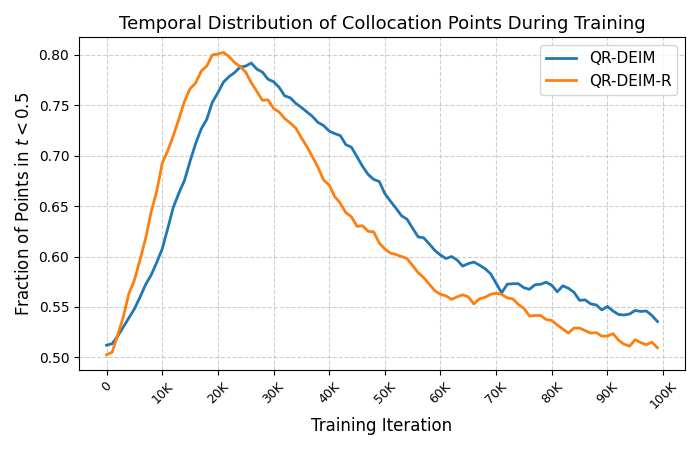}
    \caption{Fraction of collocation points in the first half of the time domain over the course of training for the QR-DEIM (blue) and QR-DEIM-R (orange) sampling methods for the wave equation.\label{fig:wave-temporal}}
\end{figure}

Figures~\ref{fig:point-distributions} and~\ref{fig:point-distributions-rand} (left columns) visually illustrate how our QR-DEIM and QR-DEIM-R methods adaptively update the training set for the wave equation. We first note that both sets of training points look similar, indicating that the QR-DEIM-R method is able to accurately approximate the behavior of its full QR-DEIM-based counterpart. Next, we observe that in both cases, the distribution of the collocation points appears to be heavily concentrated in the first half of the time interval ($t < 0.5$) between iterations 20,000 to 30,000. Figure~\ref{fig:wave-temporal} quantifies this trend by plotting the fraction of collocation points in the first half of the time domain over the course of training. Both QR-DEIM-based methods concentrate a majority of their points (over $75\%$) in the early-time region. As training progresses, the point distribution gradually shifts toward later times, suggesting that our method implicitly learns the temporal evolution of the solution.

This behavior highlights an important strength of our method: QR-DEIM and QR-DEIM-R appear to adaptively allocate training points in a way that tracks the temporal evolution of the solution. In the next example, we demonstrate how this behavior becomes even more pronounced for convection-dominated problems.

\subsection{Convection Equation}
We next test our approach on the convection equation stated below:
\begin{align*}
\begin{split}
    \frac{\partial u}{\partial t} + 20 \frac{\partial u}{\partial x}&= 0, \quad \mbox{in}\quad (-1, 1) \times (0, 1], \\
    u(-1, t) &= u(1, t), \quad \forall t \in [0, 1], \\  
    u(x, 0) &= \sin(\pi x), \quad \forall x \in (-1, 1). 
\end{split}
\end{align*}

This problem involves a sharp rightward transport of the initial sine profile at high velocity, with periodic boundary conditions ensuring continuity across domain boundaries. Such convection-dominated PDEs are known to pose significant challenges for standard PINN training, which often struggles to accurately resolve advective features and maintain temporal coherence \cite{krishnapriyan2021characterizing, mojgani2023kolmogorov, pmlr-v202-daw23a}.

We use this example to evaluate how well our QR-DEIM-based sampling strategies adapt to sharp transport features. As shown in Table~\ref{tab:results-accuracy}, both QR-DEIM and QR-DEIM-R achieve significantly lower relative errors compared to the other fixed and adaptive collocation point sampling strategies. Notably, both methods produce relative errors that are up to two orders of magnitude smaller than those achieved by existing approaches, highlighting the effectiveness of our adaptive strategy in capturing advective dynamics.

\begin{figure}[ht!]
    \centering
    \includegraphics[width=\linewidth]{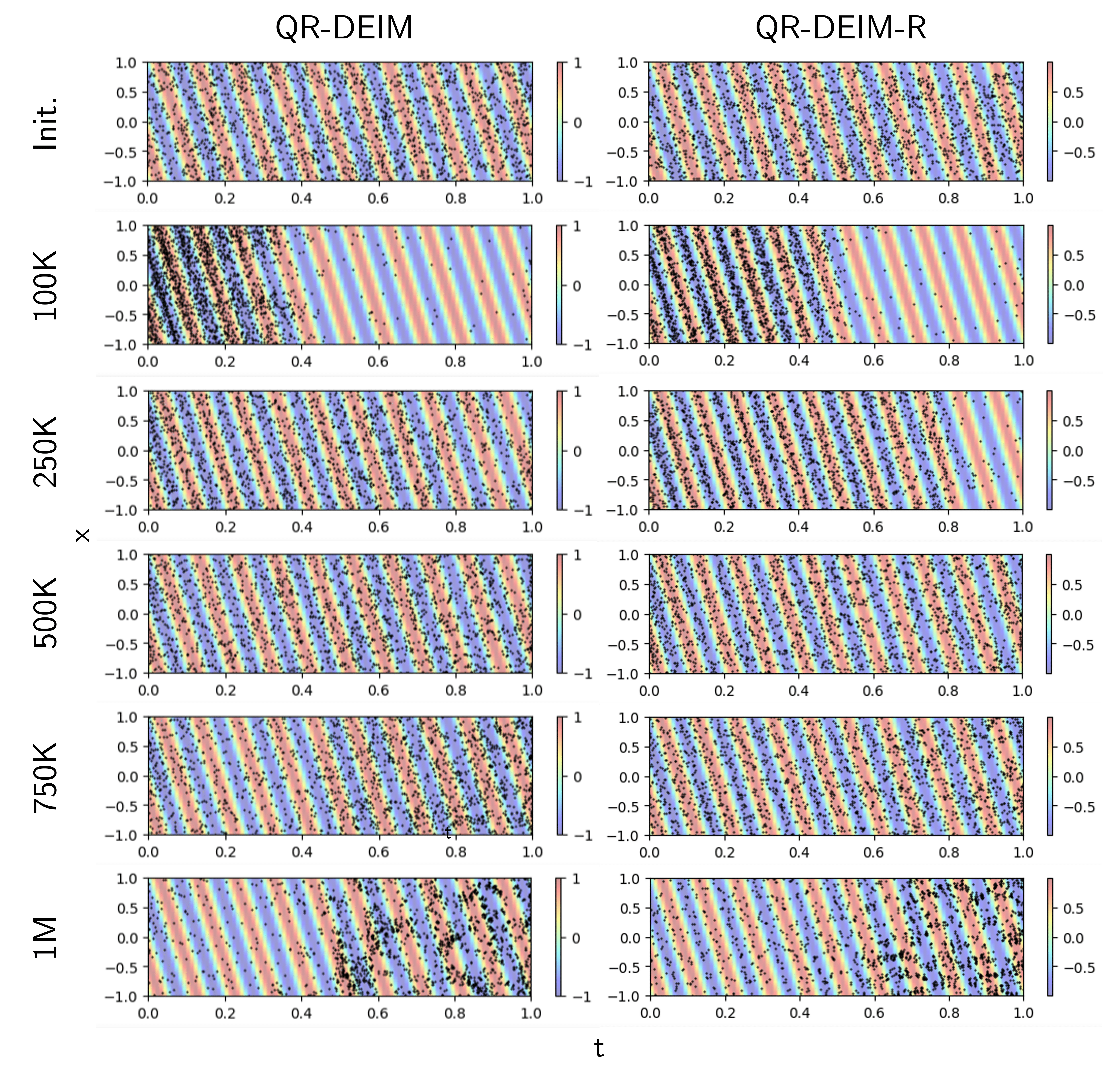}
    \caption{Evolution of points for the QR-DEIM (left) and QR-DEIM-R (right) methods for the convection equation. \label{fig:convection-points}}
\end{figure}

\begin{figure}[ht!]
    \centering
    \includegraphics[width=0.85\linewidth]{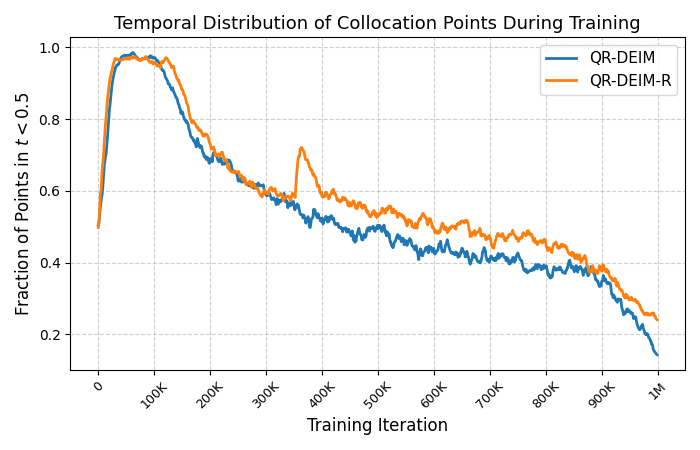}
    \caption{Fraction of collocation points in the first half of the time domain over the course of training for the QR-DEIM (blue) and QR-DEIM-R (orange) sampling methods for the convection equation.\label{fig:convection-temporal}}
\end{figure}

Like with the wave equation, we observe that both of our QR-DEIM-based methods concentrate points in the earlier portion of the time domain. Figure~\ref{fig:convection-points} illustrates this behavior for the QR-DEIM (left) and QR-DEIM-R (right) methods. At iteration 100{,}000, both methods place the majority of collocation points in the first half of the temporal domain. As training progresses, the point distributions gradually shift forward in time, becoming more balanced at iteration 250{,}000 and ultimately concentrating in the latter half of the domain by iteration 750,000. Figure~\ref{fig:convection-temporal} quantifies this trend by plotting the fraction of collocation points in the first half of the temporal domain throughout training. For both methods, we observe an early bias toward the initial time interval, where nearly all points lie in the first half of the domain for the first 200,000 iterations. This is followed by a gradual reallocation of points and, ultimately, a sharp concentration in the latter half of the domain as training approaches the final iteration.

This temporal evolution of the collocation point distribution suggests that our QR-DEIM-based methods implicitly learn to track the characteristic propagation of the solution. Rather than attempting to resolve the entire spatiotemporal domain uniformly, our approach adaptively reallocates collocation points in a temporally progressive manner, initially focusing on early-time dynamics and gradually shifting emphasis toward later times as training advances. This behavior reflects the motivation behind explicit time-marching and domain decomposition strategies proposed in prior work \cite{penwarden2023unified, jagtap2020extended, zhao2021solving}, where the temporal domain is manually partitioned and either trained in stages or assigned to specialized subnetworks. In contrast, our QR-DEIM-based approach requires no manual intervention or architectural changes. Instead, it discovers a temporally adaptive sampling strategy directly from the training dynamics and the evolving residual structure.

\subsection{Allen-Cahn Equation}
To test our approach on a nonlinear PDE, we consider the Allen-Cahn equation stated below:
\begin{align*}
\begin{split}
    \frac{\partial u}{\partial t} &= 0.0001\frac{\partial^2u}{\partial x^2} + 5(u^3 - u),\quad \mbox{in}\quad  (-1, 1) \times (0, 1], \\
    u(x, 0) &= x^2\cos(\pi x),\quad  \forall x \in (-1, 1),\\
    u(-1, t) &= u(1, t) = -1, \quad \forall t \in [0, 1].
\end{split}
\end{align*}
This equation models phase separation dynamics in binary mixtures and features nonlinear reaction terms that generate sharp phase boundaries when the diffusion coefficient is small. In our setup, the small diffusion parameter (i.e., 0.0001) leads to sharp phase boundaries that evolve slowly over time, resulting in highly non-uniform spatial features.

Table~\ref{tab:results-accuracy} presents the mean and standard deviation of the relative $\ell_2$ error for the Allen–Cahn equation using various collocation point sampling strategies. As with the wave and convection benchmarks, both QR-DEIM and QR-DEIM-R achieve lower average relative errors compared to the other methods. Figures~\ref{fig:point-distributions} and~\ref{fig:point-distributions-rand} (central columns) visually illustrate the evolution of the training collocation points for the QR-DEIM and QR-DEIM-R methods, respectively. Unlike the wave and convection problems, where the QR-DEIM-based methods tracked temporal dynamics by reallocating points over time, the Allen–Cahn equation elicits a different form of adaptive behavior. Here, the solution features localized high-gradient regions where sharp phase transitions occur. Our QR-DEIM and QR-DEIM-R methods naturally concentrate collocation points around these steep interface regions without any prior knowledge of the phase boundaries or the underlying PDE structure. This spatial selectivity highlights the ability of our method to discover and resolve critical features based solely on the evolving residual structure.

These findings underscore another flexibility of QR-DEIM-based sampling: it reacts to the underlying solution features in a data-driven manner. This adaptability appears to be valuable in nonlinear PDEs like Allen–Cahn. To further demonstrate this, we next consider Burgers' equation. This nonlinear conservation law introduces both advective and diffusive behavior and can lead to the formation of shocks depending on the initial conditions.

\subsection{Burgers' Equation}
Finally, we consider Burgers' equation given by
\begin{align*}
\begin{split}
    \frac{\partial u}{\partial t} &= \frac{0.01}{\pi} \frac{\partial^2u}{\partial x^2} - u\frac{\partial u}{\partial x},\quad \mbox{in}\quad (-1, 1) \times (0, 1], \\
    u(x, 0) &= -\sin(\pi x), \quad \forall x \in (-1, 1),\\
    u(-1, t) &= u(1, t) = 0, \quad \forall t \in [0, 1].
\end{split}
\end{align*}
This equation serves as a classic benchmark for nonlinear PDEs with both advective and diffusive components. The chosen initial condition evolves into a traveling wave that steepens over time, forming a sharp front as a result of nonlinear convection. This makes Burgers’ equation particularly challenging for PINNs, which often struggle to resolve sharp gradients or emergent shocks without excessively fine sampling.

\begin{figure}
    \centering
    \includegraphics[width=\textwidth]{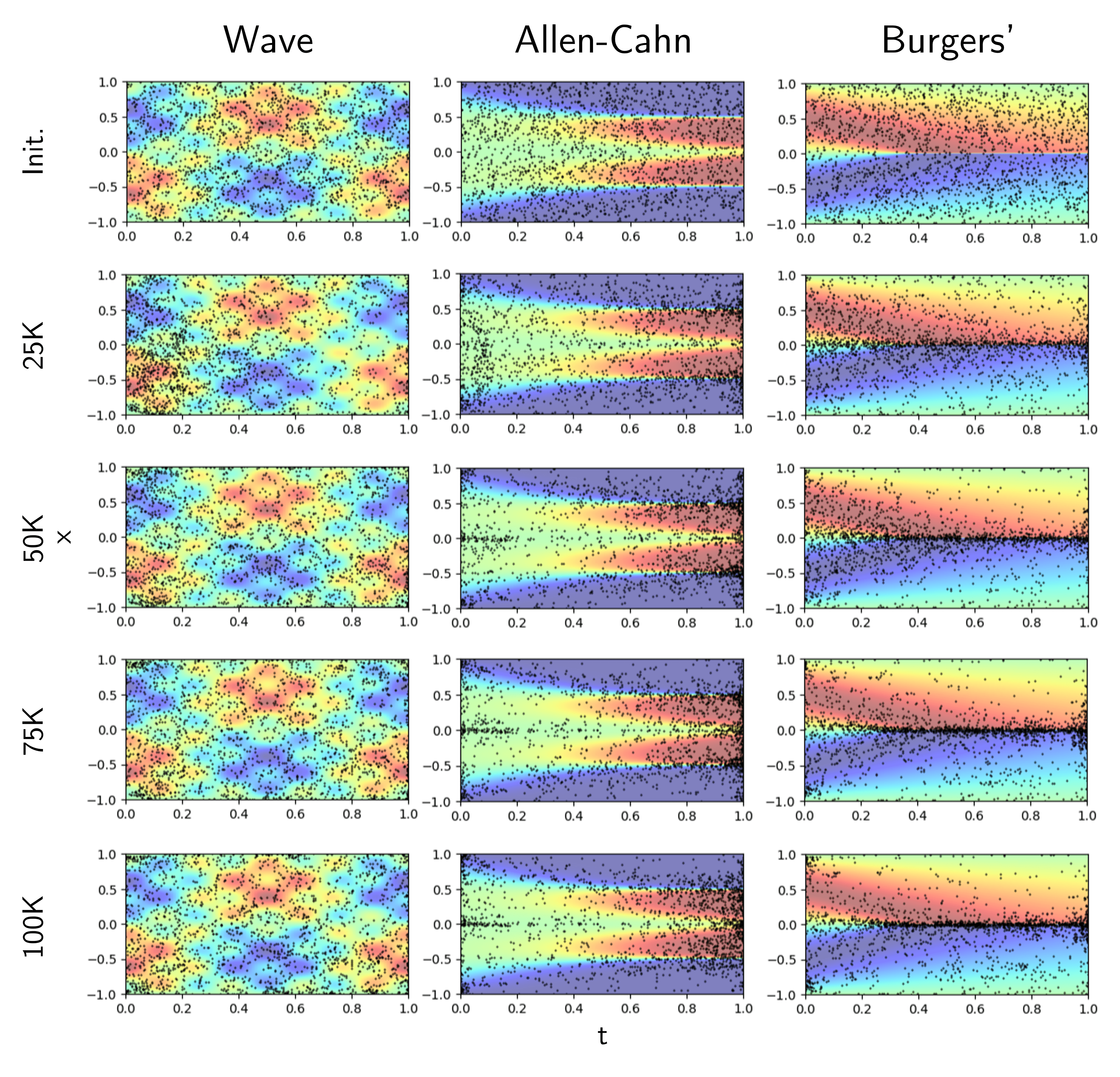}
\caption{Training collocation points (black dots) overlaid on top of the true solution for the wave (left), Allen-Cahn (center), and Burgers' (right) equations. From top to bottom, we plot the initial training set and the subsequent training sets after 25,000, 50,000, 75,000, and 100,000 training iterations for our QR-DEIM-based approach. \label{fig:point-distributions}}
\end{figure}

\begin{figure}
    \centering
    \includegraphics[width=\textwidth]{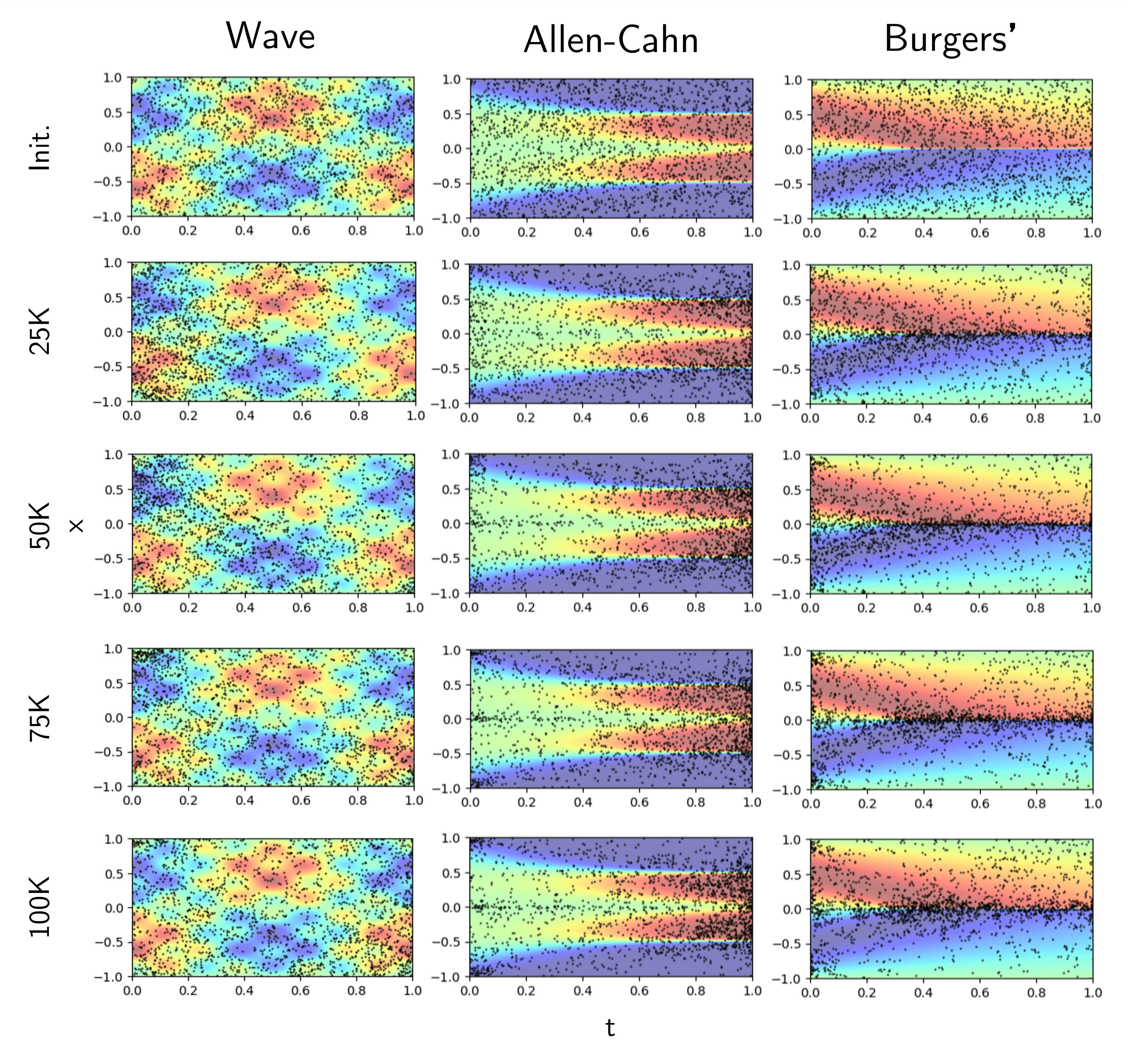}
\caption{Training collocation points (black dots) overlaid on top of the true solution for the wave (left), Allen-Cahn (center), and Burgers' (right) equations. From top to bottom, we plot the initial training set and the subsequent training sets after 25,000, 50,000, 75,000, and 100,000 training iterations for our randomized QR-DEIM approach. \label{fig:point-distributions-rand}}
\end{figure}

Table~\ref{tab:results-accuracy} presents the mean and standard deviation of the relative $\ell_2$ errors for Burgers’ equation using different sampling strategies. As with the previous benchmarks, the QR-DEIM and QR-DEIM-R methods yield lower errors than both fixed and other adaptive sampling baselines. Figures~\ref{fig:point-distributions} and~\ref{fig:point-distributions-rand} (right columns) visually illustrate the evolution of the training collocation points for the QR-DEIM and QR-DEIM-R methods, respectively. Again, we observe that, unlike the wave and convection equations, the QR-DEIM-based methods applied to Burgers’ equation exhibit a distinctly spatial adaptation. As the shock forms and sharpens near the center of the spatial domain (i.e., $x=0$), both QR-DEIM and QR-DEIM-R increasingly concentrate training points in this high-gradient region.

This spatial focusing effect is critical for accurately resolving steep nonlinear features, which standard uniform sampling schemes tend to undersample without large, dense training sets. By directing collocation points toward the shock, our methods achieve lower errors while maintaining a reasonable training set size. These results further reinforce the versatility of QR-DEIM and QR-DEIM-R: whether the solution dynamics are dominated by temporal transport, localized spatial transitions, or nonlinear shocks, the adaptive sampling strategy naturally reorients its focus based on the evolving structure of the residuals.

Together with the previous results, this final example illustrates that QR-DEIM-based sampling provides a principled and generalizable mechanism for resolving challenging features across a broad class of PDEs.

\begin{table}[ht!]
\centering
\resizebox{\textwidth}{!}{%
\setlength{\tabcolsep}{7.5pt}
\renewcommand{\arraystretch}{1.5}
\begin{tabular}{lllll}
\hline
Method & Wave & Convection & Allen-Cahn & Burgers' \\ \hline
\rowcolor[HTML]{EFEFEF} Random & 1.17e-00 (5.49e-01) & 1.06e-00 (4.60e-02) & 2.21e-02 (2.97e-02) & 2.84e-01 (1.65e-01) \\
\rowcolor[HTML]{FFFFFF} Hammersly & 3.96e-01 (4.24e-01) & 1.03e-00 (3.35e-02) & 6.18e-03 (1.57e-03) & 2.66e-01 (1.86e-01) \\
\rowcolor[HTML]{EFEFEF} Random-R & 1.80e-01 (9.65e-02) & 1.02e-00 (1.44e-02) & 9.00e-03 (1.00e-03) & 2.34e-02 (3.57e-02) \\ \hline
\rowcolor[HTML]{FFFFFF} RAR-G & 1.50e-01 (1.24e-01) & 1.35e-00 (1.48e-01) & 5.14e-03 (9.35e-04) & 3.12e-03 (2.84e-03) \\
\rowcolor[HTML]{EFEFEF} RAR-D & 9.68e-02 (8.09e-02) & 1.35e-00 (1.35e-01) & 5.53e-03 (9.63e-04) & 7.31e-04 (1.78e-04) \\
\rowcolor[HTML]{FFFFFF} RAD & 5.16e-02 (5.31e-02) & 8.07e-01 (2.09e-02) & 8.61e-03 (2.70e-03) & 1.23e-03 (4.57e-04) \\
\rowcolor[HTML]{EFEFEF} R3 & 2.25e-01 (2.39e-02) & 8.44e-01 (1.91e-02) & 1.83e-02 (2.26e-03) & 1.77e-03 (4.43e-04) \\
\rowcolor[HTML]{FFFFFF} PINNACLE & 4.96e-01 (3.31e-01) & 1.29e-00 (2.50e-01) & 3.21e-02 (3.39e-02) & 9.60e-03 (4.13e-03) \\
\rowcolor[HTML]{EFEFEF} QR-DEIM & \cellcolor{yellow}9.44e-03 (2.49e-03) & \cellcolor{yellow}6.94e-02 (4.29e-02) & 4.05e-03 (1.67e-03) & 5.96e-04 (7.07e-05) \\
\rowcolor[HTML]{FFFFFF} QR-DEIM-R & 9.96e-03 (3.47e-03) & 7.02e-02 (1.88e-02) & \cellcolor{yellow}3.85e-03 (1.87e-03) & \cellcolor{yellow}5.06e-04 (5.69e-05) \\ \hline
\end{tabular}%
}
\caption{Mean and standard deviation of the relative $\ell_2$ errors for ten simulations for each method and benchmark problem. The lowest means for each problem are highlighted in yellow.}
\label{tab:results-accuracy}
\end{table}

\subsection{Ablation Studies}
To evaluate the robustness of our proposed sampling strategies, we conduct an ablation study on the wave equation benchmark by varying key hyperparameters for both QR-DEIM and QR-DEIM-R. We begin by analyzing the effect of the energy threshold $\varepsilon$ in the full QR-DEIM method. This parameter determines how much of the residual snapshot energy must be captured by the truncated SVD, thereby controlling the number of new collocation points selected at each update. Lower values of $\varepsilon$ retain more singular vectors and thus result in a larger set of new points. Table~\ref{tab:vary-epsilon} presents the average relative $\ell_2$ errors over ten runs for various threshold values. While the default setting of $\varepsilon$ equal to 0.005 yields the lowest average error, we observe that varying $\varepsilon$ across a wide range, from 0.1 to 0.001, has a minor effect on the method’s overall accuracy, indicating that QR-DEIM is relatively insensitive to this parameter in this benchmark.

Similarly, we see that the QR-DEIM-R method is relatively insensitive to the selected rank $k$, which directly determines the rank of the randomized SVD approximation of the residual snapshot matrix. In other words, this parameter governs the dimensionality of the subspace used to approximate the residual structure and hence affects the quality of the selected collocation points. As shown in Table~\ref{tab:vary-rank-k}, varying $k$ over a reasonable range (i.e., from 50 to 250) results in only modest fluctuations in the average relative $\ell_2$ error. The lowest error is achieved with the default setting of the rank $k$ equal to 100, but the method remains robust across the tested values. We also note that, although not shown here, the oversampling parameter used in the randomized QR pivot selection has a minimal effect on the relative error. These results suggest that QR-DEIM-R, like its full QR-DEIM counterpart, is not overly sensitive to its rank parameter for the wave equation.

Finally, we examine the impact of the number of snapshot points $N_{\text{snapshot}}$ and the snapshot period $P$ on both QR-DEIM and QR-DEIM-R. These two parameters govern how many snapshot points we can choose from and how frequently we update the set of training collocation points. As shown in Tables~\ref{tab:vary-snapshots} and~\ref{tab:vary-snapshots-rand}, both methods exhibit reasonable robustness across a wide range of snapshot configurations. For QR-DEIM, the combination of $N_{\text{snapshot}}$ equal to 1,000 and $P$ equal to 500 yields the best average accuracy, while for QR-DEIM-R, the best accuracy is achieved with the same snapshot count but a slightly larger period of $P$ equal to 1,000. In both cases, increasing the snapshot frequency (i.e., smaller $P$) provides minor improvements in accuracy, but the variations remain relatively insignificant.

We observe similar patterns of robustness across the other benchmark problems presented in this work. In each case, the QR-DEIM and QR-DEIM-R methods consistently maintain low relative $\ell_2$ errors across a broad range of hyperparameter values, suggesting that their performance is not overly sensitive to tuning. However, while these initial results are encouraging, we emphasize that further investigation is needed to fully understand the effects of these parameters across a wider variety of PDEs, especially those with more complex solution structures or in higher dimensions. As such, we refrain from prescribing any single best setting and instead highlight the practical flexibility of both methods in adapting to different problems with minimal hyperparameter adjustment.

\begin{table}[ht!]
\centering
\begin{minipage}{0.48\textwidth}
\centering
\resizebox{\textwidth}{!}{%
\setlength{\tabcolsep}{7.5pt}
\renewcommand{\arraystretch}{1.2}
\begin{tabular}{ll}
\hline
Energy Threshold $\varepsilon$ & Relative $\ell_2$ Error \\ \hline
\rowcolor[HTML]{EFEFEF} 
0.1   & 3.81e-02 (4.02e-02) \\
0.05  & 2.20e-02 (1.17e-02) \\
\rowcolor[HTML]{EFEFEF} 
0.01  & 1.38e-02 (2.33e-03) \\
0.005 & 9.87e-03 (3.23e-03) \\
\rowcolor[HTML]{EFEFEF} 
0.001 & 3.39e-02 (1.36e-02) \\ \hline
\end{tabular}%
}
\caption{Relative $\ell_2$ error for varying energy thresholds $\varepsilon$ using the QR-DEIM method. All other parameters are set at their default values. Mean and standard deviation are computed over ten runs.}
\label{tab:vary-epsilon}
\end{minipage}
\hfill
\begin{minipage}{0.48\textwidth}
\centering
\resizebox{\textwidth}{!}{%
\setlength{\tabcolsep}{9.5pt}
\renewcommand{\arraystretch}{1.2}
\begin{tabular}{ll}
\hline
Selected Rank $k$ & Relative $\ell_2$ Error \\ \hline
\rowcolor[HTML]{EFEFEF} 
50  & 1.46e-02 (4.51e-03) \\
100 & 9.96e-03 (3.47e-03) \\
\rowcolor[HTML]{EFEFEF} 
150 & 3.23e-02 (2.26e-02) \\
200 & 2.17e-02 (1.43e-02) \\
\rowcolor[HTML]{EFEFEF} 
250 & 1.67e-02 (9.18e-03) \\ \hline
\end{tabular}%
}
\caption{Relative $\ell_2$ error for varying rank $k$ using the QR-DEIM-R method. All other parameters are set at their default values. Mean and standard deviation are computed over ten runs.}
\label{tab:vary-rank-k}
\end{minipage}
\end{table}

\begin{table}[ht!]
\centering
\resizebox{\textwidth}{!}{%
\setlength{\tabcolsep}{9.5pt}
\renewcommand{\arraystretch}{1.5}
\begin{tabular}{ll|lll}
\hline
\multicolumn{2}{l|}{} & \multicolumn{3}{c}{Snapshot Period $P$} \\ \cline{3-5} 
\multicolumn{2}{l|}{\multirow{-2}{*}{}} & 500 & 1,000 & 2,000 \\ \hline
\multicolumn{1}{l|}{} & 500 & \cellcolor[HTML]{EFEFEF}8.07e-03 (3.71e-03) & \cellcolor[HTML]{EFEFEF}1.79e-02 (9.94e-03) & \cellcolor[HTML]{EFEFEF}2.46e-02 (1.42e-02) \\
\multicolumn{1}{l|}{} & 1,000 & 7.59e-03 (2.33e-03) & 9.44e-03 (2.49e-03) & 1.75e-02 (8.15e-03) \\
\multicolumn{1}{l|}{\multirow{-3}{*}{\begin{tabular}[c]{@{}l@{}}Num. Snapshot\\ Points $N_{\text{snapshot}}$\end{tabular}}} & 2,000 & \cellcolor[HTML]{EFEFEF}8.77e-03 (3.86e-03) & \cellcolor[HTML]{EFEFEF}9.69e-03 (4.05e-03) & \cellcolor[HTML]{EFEFEF}1.54e-02 (3.21e-03) \\ \hline
\end{tabular}%
}
\caption{Relative $\ell_2$ error for varying numbers of snapshot points $N_{\text{snapshot}}$ and snapshot periods $P$ for the QR-DEIM method applied to the wave equation. All other parameters are set at their default values. Mean and standard deviation values are computed over ten runs.}
\label{tab:vary-snapshots}
\end{table}

\begin{table}[ht!]
\centering
\resizebox{\textwidth}{!}{%
\setlength{\tabcolsep}{9.5pt}
\renewcommand{\arraystretch}{1.5}
\begin{tabular}{ll|lll}
\hline
\multicolumn{2}{l|}{} & \multicolumn{3}{c}{Snapshot Period $P$} \\ \cline{3-5} 
\multicolumn{2}{l|}{\multirow{-2}{*}{}} & 500 & 1,000 & 2,000 \\ \hline
\multicolumn{1}{l|}{} & 500 & \cellcolor[HTML]{EFEFEF}1.89e-02 (2.54e-03) & \cellcolor[HTML]{EFEFEF}3.27e-02 (3.17e-02) & \cellcolor[HTML]{EFEFEF}2.60e-02 (1.01e-02) \\
\multicolumn{1}{l|}{} & 1,000 & 1.08e-02 (2.85e-03) & 9.96e-03 (3.47e-03) & 2.28e-02 (5.85e-03) \\
\multicolumn{1}{l|}{\multirow{-3}{*}{\begin{tabular}[c]{@{}l@{}}Num. Snapshot\\ Points $N_{\text{snapshot}}$\end{tabular}}} & 2,000 & \cellcolor[HTML]{EFEFEF}1.17e-02 (3.74e-03) & \cellcolor[HTML]{EFEFEF}1.58e-02 (7.38e-03) & \cellcolor[HTML]{EFEFEF}2.25e-02 (7.14e-03) \\ \hline
\end{tabular}%
}
\caption{Relative $\ell_2$ error for varying numbers of snapshot points $N_{\text{snapshot}}$ and snapshot periods $P$ for the QR-DEIM-R method applied to the wave equation. All other parameters are set at their default values. Mean and standard deviation values are computed over ten runs.}
\label{tab:vary-snapshots-rand}
\end{table}

\subsection{Computational Costs \& Considerations}
While the QR-DEIM and QR-DEIM-R methods yield notable accuracy improvements through adaptive sampling, they introduce a non-negligible computational overhead compared to fixed collocation strategies. This overhead arises not from the update algorithms themselves, which are relatively efficient, but from the repeated computation of residuals over a potentially large snapshot set. For example, on the wave equation with $N_{\text{snapshot}} = P = 1,000$, a fixed sampling strategy takes approximately 9 seconds per 1,000 training iterations on our hardware. In contrast, the QR-DEIM and QR-DEIM-R methods increase this to around 11 seconds, with the bulk of the additional time spent constructing the residual snapshot matrix. The update steps are comparatively inexpensive: a QR-DEIM update takes roughly 0.2 seconds for a $1000 \times 1000$ residual matrix, while the QR-DEIM-R update, benefiting from randomized numerical methods, takes just 0.02 seconds.

\begin{figure}
    \centering
    \includegraphics[width=\linewidth]{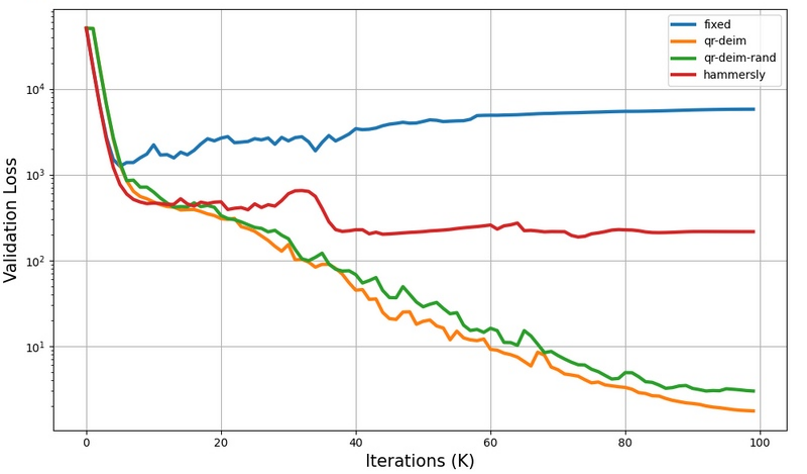}
    \caption{Average validation loss curves for the wave equation for QR-DEIM, QR-DEIM-R, and two fixed sampling baselines (uniform random and Hammersly).\label{fig:wave-val-loss}}
\end{figure}

Despite the additional computational overhead introduced by our methods, the improved sampling efficiency of the QR-DEIM and QR-DEIM-R methods leads to lower validation loss values in fewer iterations. Figure~\ref{fig:wave-val-loss} shows the average validation loss curves for the wave equation for both QR-DEIM variants and two fixed methods (uniform random and Hammersly sampling). While the fixed methods plateau early and exhibit slower convergence, both QR-DEIM and QR-DEIM-R steadily reduce the validation loss throughout training, ultimately reaching significantly lower error levels in fewer iterations. In our view, the ability to obtain more accurate solutions with fewer training steps more than compensates for the modest per-iteration cost, underscoring the effectiveness of our methods in accelerating PINN training while maintaining reasonable computational demands.

While both QR-DEIM and QR-DEIM-R are effective for the 1D benchmarks considered in this work, their computational profiles differ in important ways. In particular, the randomized variant, QR-DEIM-R, offers substantial advantages when working with larger snapshot sets or less frequent updates, conditions that might be useful in higher-dimensional problems or more complex PDEs. Although full QR-DEIM can be efficiently computed for modestly sized snapshot matrices using commonly available, highly optimized numerical linear algebra libraries, its computational cost scales cubically with the number of snapshot points. In contrast, QR-DEIM-R leverages randomized methods to achieve a significantly cheaper update step, with cost linear in the number of retained singular vectors and matrix dimensions. As such, QR-DEIM-R is better suited for scenarios where memory or runtime constraints limit the feasibility of full QR-based updates. In the problems studied here, both methods yield similar accuracy and convergence patterns, but the lower update cost of QR-DEIM-R may become increasingly important as problem complexity grows.

Compared to other adaptive sampling strategies in the PINN literature, the computational cost of our methods lies roughly in the middle of the spectrum. QR-DEIM and QR-DEIM-R are more expensive than R3, which avoids residual snapshots altogether and instead reuses the residual computation needed for evaluating the loss function. On the other hand, we observed that the computational costs of our QR-DEIM-based methods are comparable to those of RAD, RAR-D, and RAR-G, all of which require evaluating the residual on a large, dense candidate pool at each update step. Finally, our methods are computationally less expensive than PINNACLE, which involves additional eigendecomposition steps and kernel-based operations. Overall, QR-DEIM and QR-DEIM-R provide a practical trade-off between computational cost and sampling effectiveness. They are well-suited for scenarios where moderate overhead is acceptable in exchange for improved convergence and accuracy.

\section{Conclusions}
In this paper, we presented two adaptive collocation point selection strategies for PINNs, both rooted in the QR-DEIM algorithm. The first approach utilizes the standard QR-DEIM algorithm, whereas the second incorporates a randomized variant to enhance scalability. We assessed both methods across four benchmark problems - the wave, convection, Allen–Cahn, and Burgers’ equations - and compared their performance against a broad spectrum of fixed and adaptive sampling strategies. Our findings indicate that both methods consistently yield lower errors than existing techniques, establishing their efficacy as robust and efficient tools for adaptive sampling within PINNs.

Looking ahead, we intend to extend these methods to higher-dimensional or coupled PDEs. Additionally, future work will investigate how QR-DEIM-based sampling methods perform in conjunction with more advanced training techniques. However, based on our current training framework, the results imply that QR-based adaptive sampling is a promising avenue for enhancing the reliability and efficiency of PINNs in complex scientific computing applications.

\bibliographystyle{ieeetr}
\bibliography{sources}
\end{document}